\documentclass[10pt,conference, compsocconf]{IEEEtran}
\IEEEoverridecommandlockouts
\usepackage{cite}
\usepackage{amsmath,amssymb,amsfonts}
\usepackage{algorithmic}
\usepackage{graphicx}
\usepackage{textcomp}
\usepackage{xcolor}
\usepackage{hyperref}

\usepackage[capitalise]{cleveref}

\usepackage{flushend}

\def\BibTeX{{\rm B\kern-.05em{\sc i\kern-.025em b}\kern-.08em
    T\kern-.1667em\lower.7ex\hbox{E}\kern-.125emX}}
\begin{document}

\title{PresSim: An End-to-end Framework for Dynamic Ground Pressure Profile Generation from Monocular Videos Using Physics-based 3D Simulation}

\author{\IEEEauthorblockN{
Lala Shakti Swarup Ray\IEEEauthorrefmark{1}\IEEEauthorrefmark{2}, Bo Zhou\IEEEauthorrefmark{1}\IEEEauthorrefmark{2}, Sungho Suh\IEEEauthorrefmark{1}\IEEEauthorrefmark{2} and Paul Lukowicz\IEEEauthorrefmark{1}\IEEEauthorrefmark{2}}
\IEEEauthorblockA{\IEEEauthorrefmark{1}German Research Center for Artificial Intelligence (DFKI), 67663 Kaiserslautern, Germany}
\IEEEauthorblockA{\IEEEauthorrefmark{2}Department of Computer Science, RPTU Kaiserslautern-Landau, 67663 Kaiserslautern, Germany}
\IEEEauthorblockA{Email: \{lala\_shakti\_swarup.ray, bo.zhou, sungho.suh, paul.lukowicz\}@dfki.de}
}

\maketitle
\setcounter{page}{1}

\begin{abstract}
Ground pressure exerted by the human body is a valuable source of information for human activity recognition (HAR) in unobtrusive pervasive sensing. 
While data collection from pressure sensors to develop HAR solutions requires significant resources and effort, we present a novel end-to-end framework, PresSim, to synthesize sensor data from videos of human activities to reduce such effort significantly.
PresSim adopts a 3-stage process: first, extract the 3D activity information from videos with computer vision architectures; then simulate the floor mesh deformation profiles based on the 3D activity information and gravity-included physics simulation; lastly, generate the simulated pressure sensor data with deep learning models.
We explored two approaches for the 3D activity information: inverse kinematics with mesh re-targeting, and volumetric pose and shape estimation.
We validated PresSim with an experimental setup with a monocular camera to provide input and a pressure-sensing fitness mat (80x28 spatial resolution) to provide the sensor ground truth, where nine participants performed a set of predefined yoga sequences. 
Comparing the synthesized pressure map with the pressure sensor's ground truth based on pressure shapes is quantified through an R square value of 0.948 on the binarized pressure maps and precision of activated sensing node pressure values with corrected R square value of 0.811 within areas with ground contact.
We publish our nine-hour dataset and the source code to contribute to the broader research community.
\end{abstract}


\section{Introduction}
Body dynamics and kinematics are essential for correctly understanding human activity contexts. 
The pressure exerted by the body on its surroundings because of the combination of body mass distribution and skeletal movement is one of the essential components of human body dynamics. 
It has various clinical applications that include detecting different types of disorders in humans like hemiplegia, diagnosing injuries\cite{brund2017medial}, in rehabilitation after surgeries \cite{hwang2018estimation}, analyzing human locomotion \cite{taha2016finite}, personal fitness \cite{galarza2020effect,wkasik2019changes}, and in bio-mechanics \cite{lorenzini2018synergistic}.

Using physical tactile sensing devices such as wearables and external pressure sensors to measure ground pressure exerted by the body can be time-consuming, expensive, and prone to malfunctions. 
One solution is to leverage neural networks to extract human pressure from other comparatively easier to collect sources of information, like videos. 
This method has a specific advantage, especially in prolonged monitoring and with tasks requiring a large area of operations over physical pressure sensors. 
However, It is challenging to use body pose in combination with shapes extracted from videos to estimate ground pressure data. 
Some past works used deep neural networks to estimate the ground pressure exerted by the human body from data extracted from RGB videos  \cite{scott2020image} \cite{clever2021bodypressure} but are limited to specific poses and organs of the body. 

\begin{figure*}
\begin{center}
\includegraphics[width=0.9\linewidth]{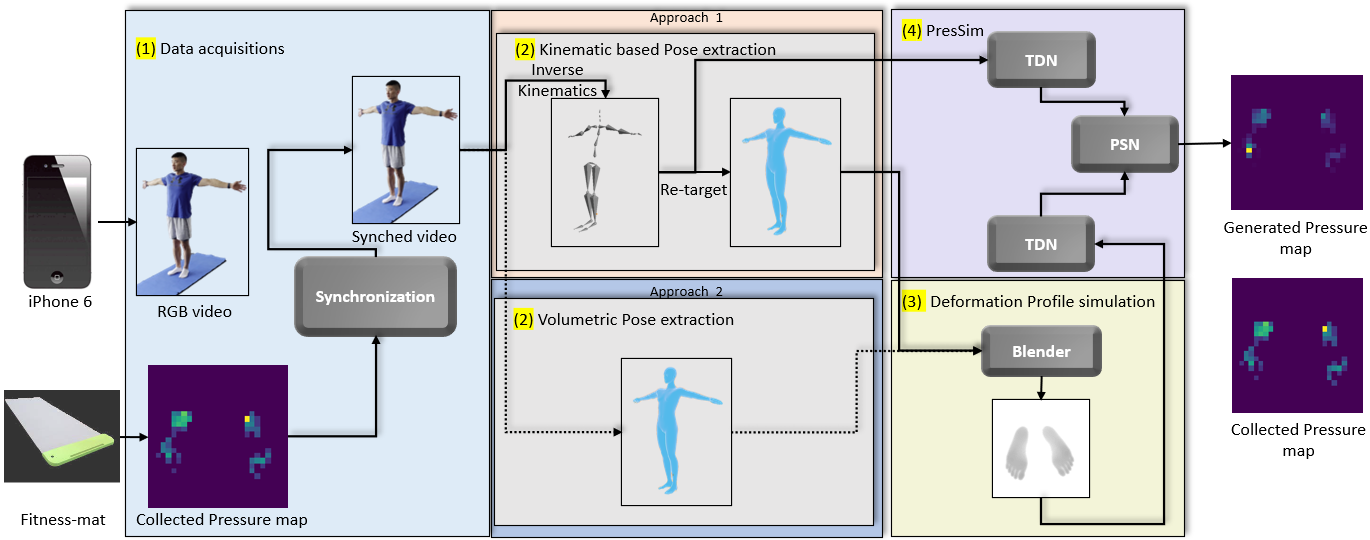}
\end{center}
   \caption{Illustration of the proposed end-to-end pipeline that consists of 1. Data Acquisition, 2. Pose Extraction, 3. Deformation Profile Simulation (inverse
kinematics with mesh re-targeting, or volumetric pose and shape estimation), and 4. Pressure Map Generation.}
\label{fig:4}
\end{figure*}

We propose a novel solution to the complex problem by combining body shape and poses and using physics-based 3D simulation to create an intermediate datatype that has both the knowledge of body shape and position. In our first approach, we adopt a kinematics-based pose estimation model to extract 3D kinematics and then use inverse kinematics to create an animated sequence. It is re-targeted to a humanoid mesh of equivalent mass and identical bone structure to simulate deformation profiles using physics-based 3D simulation. Afterward, the simulated data and extracted 3D kinematics are used to train our temporal neural network regressor to generate the synthetic pressure maps. 
Our second approach uses a volumetric model for pose estimation based on SMPL+H \cite{pavlakos2019expressive} model for more accurate deformation profiles. 

The contributions of our work include the following:
\begin{itemize}
    \item A novel approach to infer ground pressure maps exerted by the human body from monocular videos in the wild using a combination of physics-based simulations and a 3D convolutional neural network as illustrated in \cref{fig:4}.
    \item A pipeline to simulate intermediate deformation profiles from videos using either inverse kinematics, mesh re-targeting or Volumetric pose and shape estimation along with  3D physics-based simulation.
\end{itemize}

\section{Related Work}
\label{rl}


\subsection{Pose estimation}
In computer vision, pose estimation predicts a person's joint positions from an image or a video. 
Predicting 3D pose using multi-camera motion capture systems \cite{nakano2020evaluation, freemocap}  has problems since all cameras must be calibrated beforehand, and synchronization must exist between them to predict the pose data accurately. 
Another alternative is to use a monocular pose estimation model that uses the output of 2D pose estimation models like Mediapipe \cite{singh2021real}, Detectron \cite{wu2019detectron2}, and Openpose \cite{osokin2018real} to reconstruct the joint positions in 3D space as done in VideoPose3D \cite{pavllo:videopose3d:2019} and Blazepose \cite{grishchenko2022blazepose}.
We used Detectron2 and VideoPose3D for depth estimation to reconstruct the kinematic body joints in 3D space identical to the approach from Scott, et al.\cite{scott2020image}
For volumetric pose estimation, we used easymocap\cite{easymocap} that estimates SMPL+H body shape and pose from monocular videos.


\subsection{Physics simulation}

Physics-based simulation is about creating a natural world occurrence/physical phenomenon and predicting how the real-world objects would react to it by defining the state of the natural world before starting the simulation \cite{sangaraju20213d}. Then the effect of such occurrence is observed to create more robust designs. 
It is used in the aviation and automobile industry for testing prototypes, visualizing cloth simulations \cite{li2021deep}, object deformations \cite{fang2021guaranteed}, studying human bio-mechanics \cite{michaud2021biorbd}.
We used Eevee, a 3D Physics game engine inside Blender \cite{blender}, to render the contact points between a human model and ground and pressure using rigid body simulation and dynamic paint deformation similar to Clever et al.'s \cite{clever2021bodypressure} approach.

\section{Data Collection and Deformation Simulation}
\label{dc}
\subsection{Data Acquisition}
To train our model, we have collected an extensive bi-modal dataset having synchronized whole-body monocular videos and pressure map sequences using a Fitness-mat from Sensing.Tex with a \(80 \times 28\) sensor grid having a sensing area of \(560\times 1680\) \(mm \) with separate sensor area of \(12\times 16 \) \(mm \) with measuring range of 0-5000 millimeters of mercury (mmHg). 
The hardware consists of a non-elastic, non-slippery yoga mat made with Thermoplastic Polyolefins(TPE) polymers having a range of operation from 5\% – 70\% humidity and 15$ ^\circ C $  – 45 $ ^\circ C $.
The videos are collected using an 8-megapixel (\(3840\times2160\) pixels) iPhone 6 camera with Full HD \((1920\times1080p)\) resolution at 30 frames per second.
We considered nine persons of diverse ages, sex, and body characteristics while creating our dataset.
We created a continuous yoga pose sequence containing 27 poses that involve complex body movements that anyone can perform, regardless of gender or age. 
Each sequence lasts about 20 minutes, and the subjects try to follow the routine through visual and audio inputs making the total duration of the dataset around 9 hours.
Information about each data collection session is detailed in \cref{tab2}.

\begin{table}
  \begin{center}
    {\small{
\begin{tabular}{lcccc}
\hline
Subject & Mass(\(kg\)) & Height(\(cm\)) & Gender & Frames \\
\hline
1 & 74.3 & 178 & Male & 40788 \\
2 & 76.3 & 183 & Male & 40428 \\
3 & 72.2 & 178 & Male & 43895 \\
4 & 94.7 & 172 & Male & 43630 \\
5 & 60.2 & 157.5 & Female & 27932 \\
6 & 57.7 & 162 & Female & 30359 \\
7 & 57.2 & 160 & Female & 31600 \\
8 & 52.3 & 162 & Male & 31608 \\
9 & 57.3 & 159 & Female & 31023 \\
\hline
Total & - & - & - & 321263 \\
\hline
\end{tabular}
}}
\end{center}
\caption{ Data Statistics Including Subject Mass (\(kilogram\)), Height (\(centimeter\)) and Gender.}
\vspace{-20pt}
\label{tab2}
\end{table}
The pressure data is converted to arrays of size \(f \times 80 \times 28\) where \(f\) is the total number of frames, \(80 \times 28 \in N^2 \) depicts the sensor grid size. 
Output sensor data has a range from 0 to 5000 mmHg.
The collected video data has a frame rate of 30 fps, while the pressure map sequences have a frame rate of around ten fps, and both data are manually synchronized using timestamps.
For our first approach, we used Keypoint R-CNN from Detectron2, a 2D pose estimation model trained on COCO dataset\cite{lin2014microsoft} outputs 17 key points to estimate the critical points at each frame.
To reconstruct the 3D kinematics from the sequence of the key point positions, we used VideoPose3D. 
It considers the temporal data while predicting the depth of the 2D key points and is superior to other models at predicting poses from videos of activities\cite{wang2021deep}. VideoPose3D is trained on Human3.6M dataset\cite{ionescu2013human3}.
Output body kinematics is an array of sizes \(f\times17\times3 \in \mathbb{R}^3\) where f is the total number of frames which is \(Seconds\times fps(10)\). Each frame contains 17 joints with 3D (X, Y, Z) coordinates.

For our second approach, we use EasyMocap, a volumetric pose estimation model that estimates both the 3D pose and body shape \cite{dong2020motion}. 
Unlike our previous approach, this model tracks and estimates hands and legs by considering \(f\times25\times3 \in \mathbb{R}^3\) key points instead of\(f\times17\times3 \in \mathbb{R}^3\). 
The simulated intermediate data is more similar to the collected pressure maps compared to our first approach.

\begin{figure}
\begin{center}
\includegraphics[width=\linewidth]{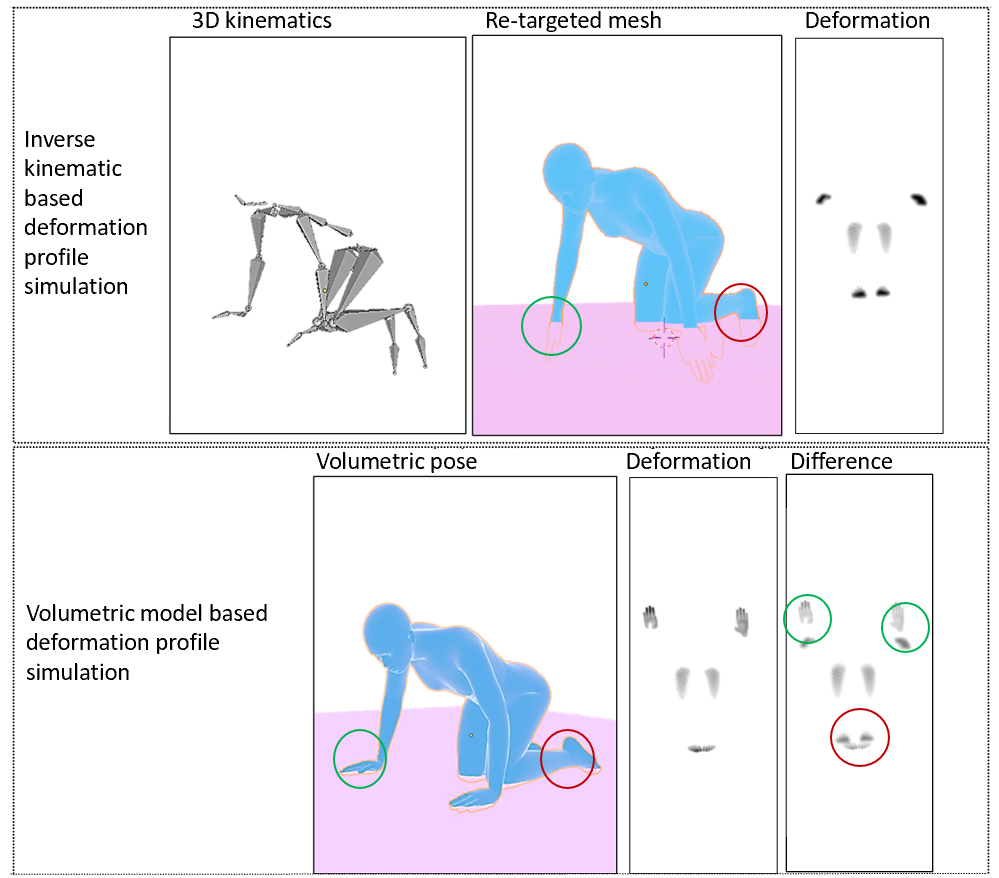}
\end{center}
\vspace{-10pt}
   \caption{Deformation profile simulation based on inverse kinematic approach and Volumetric approach: Extracted 3D pose, re-targeted 3D mesh, and corresponding deformed ground plane inside 3d simulation and difference between the simulated deformation based on two approaches.
   }
\vspace{-10pt}
\label{fig:2}
\end{figure}

\subsection{Deformation Profile Simulation}
We present a deformation profile simulation pipeline using physics simulations capable of automated creation of large datasets of deformation profiles using the collected 3D pose data and the estimated 3D humanoid mesh.
The correlation between collected pressure maps $P_{80 \times 28} \in \{0-5000\}$ and simulated deformation profiles $\bar{P}_{80 \times 28} \in \{0-255\}$ can be expressed by the following mathematical expression:
\begin{equation}
\resizebox{.9\hsize}{!}{$\left[ {\begin{array}{cccc}
    p_{1,1} & p_{1,2} & \cdots & p_{1,28}\\
    p_{2,1} & p_{2,2} & \cdots & p_{2,228}\\
    \vdots & \vdots & \ddots & \vdots\\
    p_{80,1} & p_{80,2} & \cdots & p_{80,28n}\\
  \end{array} } \right] = \alpha\left[ {\begin{array}{cccc}
    \bar{p}_{1,1} & \bar{p}_{1,2} & \cdots & \bar{p}_{1,28}\\
    \bar{p}_{2,1} & \bar{p}_{2,2} & \cdots & \bar{p}_{2,228}\\
    \vdots & \vdots & \ddots & \vdots\\
    \bar{p}_{80,1} & \bar{p}_{80,2} & \cdots & \bar{p}_{80,28n}\\
  \end{array} } \right]$}
  \label{eq:df}
\end{equation}
Where $\alpha$ is a multiplication factor constant for all pressure points at a particular frame but varies from frame to frame and needs to be estimated using neural networks.

For our first approach, we convert the extracted 3D kinematics arrays into Bio Vision Hierarchy (BVH) \cite{chung2004mcml}  motion capture format using inverse kinematics.
We used Blender \cite{blender}, an Open-source 3D software, for most of our work on 3D mesh generation and physics simulation. 
We create a 3D humanoid mesh similar to the subject's body and rig it with a 17-joint CMU skeleton identical to that from our motion capture file and re-target the motion capture file with our humanoid mesh that also preserves skeletal information like the relative distance between each joint. This information is carried on to the rigged mesh when we re-targeted it with the generated motion capture file.
Then, we converted the humanoid mesh into a rigid body and assigned it a mass equivalent to the weight of the participant used for data collection.
We created a deformable plane below the humanoid mesh and converted it to a dynamic paint canvas while the humanoid mesh acts as a dynamic paintbrush. 
Based on the point of contact between the humanoid mesh and plane are distorted up to a certain threshold but return to their original shape once the point of contact is removed.
To ensure the simulated deformation profiles do not suffer from any unintentional movement that may result in noise generation, we constrained the movement of the rigid body in X and Y directions and the rotation in all directions such that the mass of the body is the only factor that affects how the plane reacts to the human body.
We created a custom render viewpoint of size \((80\times28)\) pixels from the inbuilt Blender camera using composting nodes, identical to the sensor arrays of Fitness-mat. 

The viewpoint captures the rendered output in terms of deformation profiles. 
Based on the proximity of each vertice of the plane from the rendering viewpoint, it is represented by a number from 255 to 0, which is analogous to varying pressure points.
The output is then rendered in grey-scale PNG format for each frame using the Eevee render engine.
Afterward, OpenCV \cite{opencv_library} is used to convert the image sequences into an array of size \(f\times80\times28\).

For our second approach, we imported the extracted volumetric SMPL+H poses into Blender using the SMPL blender add-on. Then we followed the exact process as before to simulate the deformation profiles.

Representation of each intermediate step from generating motion capture file from 3D pose output to simulating deformable pressure profiles is illustrated in \cref{fig:2} based on both of our approaches. The main advantage of simulated deformation profiles by the volumetric model-based approach from the inverse kinematics approach is that the shape and position of limb terminals, as well as the overall shape of the human body, are considered in the animated mesh.

\section{PresSim: Pressure Map Synthesis}
\label{ps}

\begin{figure}
\begin{center}
\includegraphics[width=1\linewidth]{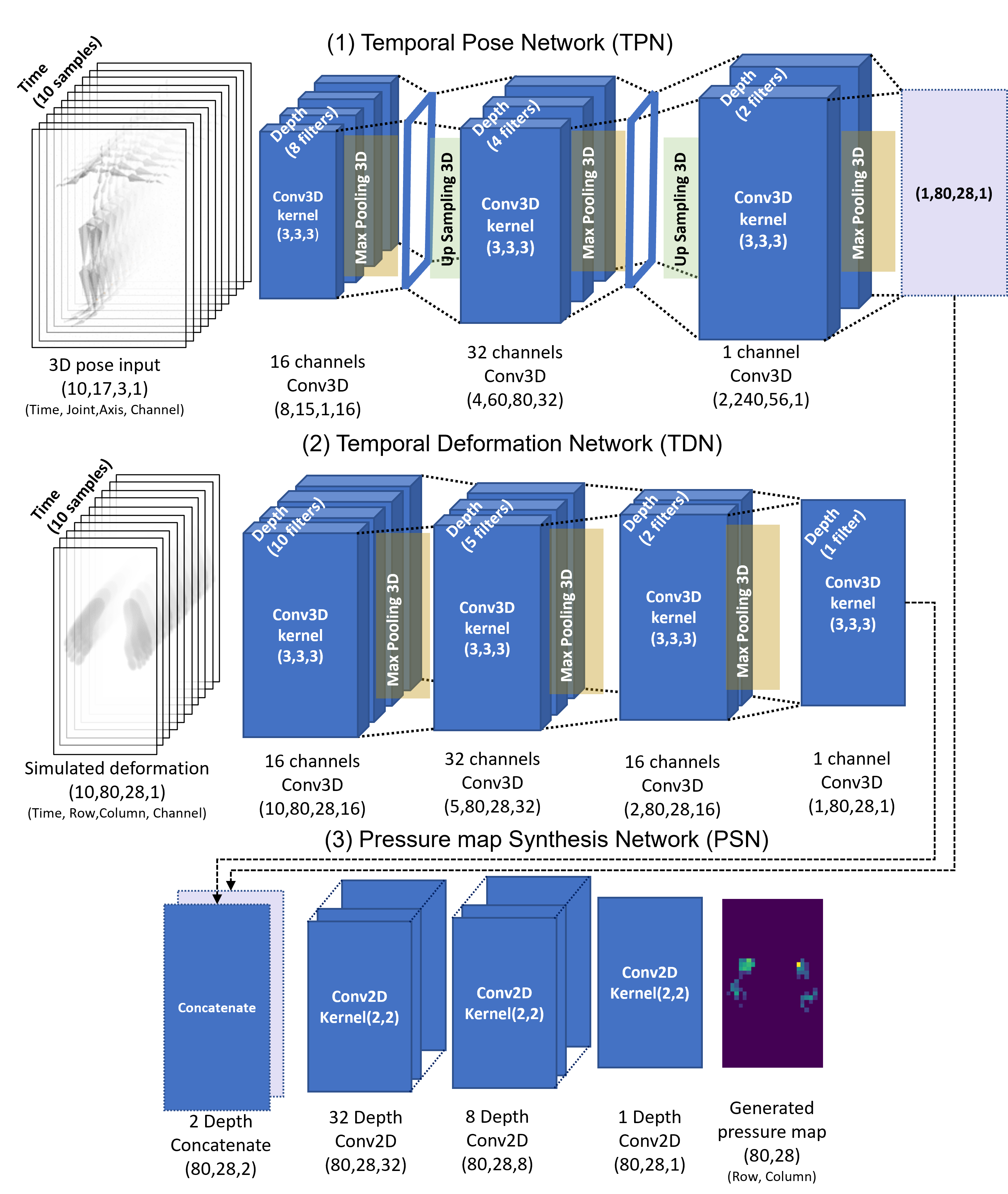}
\end{center}
   \caption{Overall PresSim Architecture. PresSim consists of three networks: Temporal Pose Network (TPN), Temporal Deformation Network (TDN), and Pressure map Synthesis Network (PSN)}
\vspace{-10pt}
\label{fig: 3}
\end{figure}

\subsection{Problem Formulation}
Our model aims to synthesize pressure sensor data directly from videos by utilizing an intermediate data type that can be simulated from the poses extracted from the videos using physics-based simulation.
It is challenging to synthesize pressure maps from 3D kinematics using deep learning techniques since the pressure maps depend heavily on the body's shape.
3D kinematics have no information about the body shape, which is essential in combination with 3D pose to generate accurate ground pressure maps.
We used a physics-based simulation system to create the intermediate data type derived from the 3D activity information extracted from the videos. 
We trained our network to learn the correlation between the kinematics and simulated deformation profiles with the corresponding body pressure map distribution.
We used a temporal 3D CNN regressor that considers ten frames (1 sec approx) of temporal pose sequence and associated deformation profiles to generate a single frame of pressure profile with the help of a sliding window.
The input for the deep learning model is ten sequential frames from \(f\times17\times3 \in \mathbb{R}^3\) array of pose data and ten frames of simulated deformation profile from \(f\times80\times28 \in \mathbb{N}^2\) array, selected by two synchronized sliding windows.
The ground truth is one frame from a \((f-10)\times80\times28 \in \mathbb{N}^2\) array of collected pressure maps.

\subsection{PresSim Architecture}

PresSim consists of three neural networks. Two 3D temporal convolutional neural networks, TPN and TDN, estimate 2D feature maps from 3D kinematics and 2D simulated deformation profiles, respectively. 
A 2D CNN called PSN is used to concatenate the output of the first two networks to synthesize 2D pressure maps.
All three networks, along with their input and output, are illustrated in \cref{fig: 3}.
\paragraph{Temporal Pose Network (TPN)}
Temporal Pose Network (TPN) is a 3D temporal CNN that takes ten frames of the extracted pose of size \(10\times17\times3\) as input and produces a 2D feature map of size \(80\times28\) as its output.
The primary purpose of TPN is to predict the average values of pressure across the horizontal axis, which is then used afterward to generate the synthesized pressure maps.
We used Adam optimizer and L2 loss for training our network. The network estimates the pressure values labeled 2D sensor positions
\{$\bar{p}_1,\bar{p}_2...,\bar{p}_N$\} where each $\bar{p}_j\in \mathbb{N}^2$ represents a 2D pressure estimate for sensor j.
We compute the loss from the square value of Euclidean error on each joint:
\begin{equation}
    L_{TPN} = \sum_{j=1}^{n} (p_n - \bar{p}_n)^2
    \label{eq:TPN}
\end{equation}
We used up-sampling layers to increase the input dimension to fit the collected pressure maps' size. We implemented a sliding window that takes ten sequential frames from the 3D array as a unit data point input to train our model. TPN has twelve layers and 15,361 parameters, of which 15,265 are trainable while 96 are non-trainable.

\paragraph{Temporal Deformation Network (TDN)}

Temporal Deformation Network (TDN) is another 3D temporal neural network regressor that takes ten frames of the simulated deformation profiles of size \(10\times80\times28\) as input and produces a 2D array of size \(80\times28\) as its output.
The primary purpose of TDN is to accurately predict the shape and relative pressure value across the predicted shapes in the 2D  feature map.
Like TPN, we used ADAM optimizer and L2 loss to train our network. The network outputs an estimate of the pressure values labeled 2D sensor positions
\{$\bar{q}_1,\bar{q}_2...,\bar{q}_N$\} where each $\bar{q}_j\in \mathbb{N}^2$ represents a 2D pressure estimate for sensor j.We compute the loss from the square value of Euclidean error on each joint:
\begin{equation}
   L_{TDN} = \sum_{j=1}^{n} (q_n - \bar{q}_n)^2
   \label{eq:TDN}
\end{equation}
A sliding window with the same clock as the sliding window used in TPN is created to take ten sequential frames from the 3D array as input as a unit datapoint to train our model. 
TDN has eleven layers consisting of 40,563 parameters, of which 40,371 are trainable while 192 are non-trainable.

\paragraph{Pressure map Synthesis Network (PSN)}

The output of the TPN, a 2D feature map of size \(80\times28\), and the TDN, a 2D feature map of identical size, are concatenated using the Pressure Map Synthesis Network (PSN), a 2D convolutional neural network. 
Similar to previous networks, we used ADAM optimizer and L2 loss for training our network. The network outputs an estimate of the pressure values labeled 2D sensor positions
\{$({\alpha \bar{p}_1 + \beta \bar{q}_1}),({\alpha \bar{p}_2 + \beta \bar{q}_2})...,({\alpha \bar{p}_N + \beta \bar{q}_N})$\} where each $\bar{p}_j, \bar{q}_j\in \mathbb{N}^2$ represents a 2D pressure estimate for sensor j derived from \cref{eq:TPN} and \cref{eq:TDN} respectively.$\alpha$ and $\beta$ are variable weights optimized for minimum sum score. We compute the loss from the square value of Euclidean error on each joint:
\begin{equation}
   L_{PSN} = \sum_{j=1}^{n} \{(\alpha |p_n - \bar{p}_n| + \beta |q_n -  \bar{q}_n|\} ^2
   \label{eq:PSN}
\end{equation}
Additionally three 2D convolutional layers are used to fit the concatenated array to the dimension of the collected pressure maps. PSN consists of 23,873 parameters (23,777 trainable). 
The combined result of both models gives us more accurate synthetic pressure maps than using only one datatype.
\begin{figure}
\begin{center}
\includegraphics[width=1\linewidth]{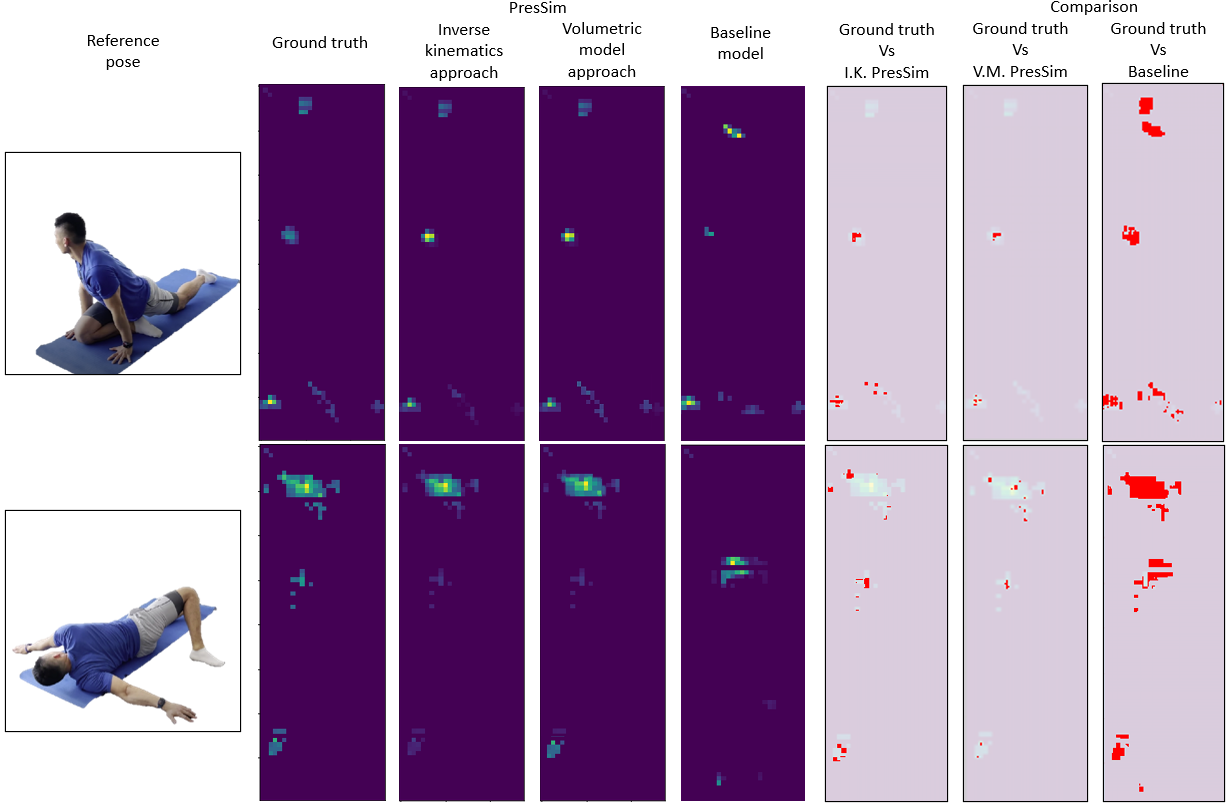}
\end{center}
   \caption{Reference video frame, ground truth collected using Fitness-mat, pressure map synthesized using PresSim based on inverse kinematic and volumetric model approach, pressure map synthesized using  baseline model and all points with different pressure value ignoring \(\pm\)10 mmHg .}
\vspace{-15pt}
\label{fig:5}
\end{figure}

\section{Experimental Results }
\label{ex}
\subsection{Implementation Details}
Our network is implemented using TensorFlow Keras 2.9.1. 
We trained our model by minimizing Mean Square Error loss using a learning rate of \(1e^{-4}\) and a batch size of 128, with Mean Absolute Error as the evaluation metric. Dropout layers of 30\% are used throughout TPN and TDN.
We use our neural network to regress the confidence value through the \(relu\) activation function. 
For PSN, a Concat layer combines the output of the first two models, and three subsequent Connv2D layers are used to fit the dimension with the pressure map array.
Our model is trained in a cluster with 128GB memory, four processors, and two Nvidia RTX A6000 graphics memories.

\subsection{Evaluation Metrics}
\label{em}
We used Mean Absolute Error as the metric to evaluate PresSim. 
For the given dataset we used to train PresSim, the range of the data is from 0 to 5000, where MAE depicts the deviation from the actual values.
To understand the correctness of the synthesized pressure maps concerning the collected sensor data, we used the R squared ($R^2$) error metric to justify the model's performance as the value ranges between 0 to 1.

For comparing the pressure profile shapes synthesized by PresSim $\bar{P}_{80 \times 28} \in \{0-5000\}$ with that of groundtruth $P_{80 \times 28} \in \{0-5000\}$, we binarized both of those arrays by translating it to $\bar{P}_{80 \times 28} \in [0,1]$ and $P_{80 \times 28} \in [0,1]$ where all values greater than zero is one.

To create a truthful comparison of the precision of each synthesized pressure node from ground truth, we calculated $R^2$ only at the point of contact.
We created a list by indexing all points in the array with non-zero pressure values and used it as a mask to get equivalent pressure data from the synthesized pressure maps and ground truth $\bar{P}_{80 \times 28} \forall \bar{p}_{i,j} \geq 1 \cap P_{80 \times 28} \forall p_{i,j} \geq 1$. 
We then calculated Root Mean Square Difference between them and used that RMSD value to calculate the corrected $R^2$  score.
\cref{fig:5} illustrates the difference between the ground truth and synthesized pressure maps using Mask - RMSD for our baseline model and PresSim trained on data simulated using both approaches.

For comparison, we trained a baseline model identical to Temporal Pose Network with some added fully connected layers only with the extracted 3D kinematics from videos.

\subsection{Qualitative Evaluation}

Regarding Mean Absolute Error, our baseline model has an average score of 19.7, while PresSim trained on inverse kinematics-based deformation profiles has an average MAE score of 10.6, and the volumetric model-based PresSim has an average of 10.4.
The $R^2$ value calculated using binarized pressure maps has an average of 0.659 for baseline, while the average value of the inverse kinematic-based PresSim approach and volumetric model-based PresSim approach is 0.909 and 0.948.
The corrected $R^2$ calculated from Mask - RMSD has an average value of 0.407 for the entire dataset for our baseline model. Our inverse kinematics-based PresSim approach has an average value of 0.8, while the average is 0.811 for our volumetric-based PresSim approach.
Both models have significant improvements compared to the baseline model.
The slight improvement in volumetric-based PresSim is because of improved deformation profile generation and no significant error in simulation because of the lack of inverse kinematics in the simulation pipeline.
All estimated average score for each model is mentioned in \cref{table3}.

\begin{table}
  \begin{center}
    {\small{
\begin{tabular}{p{2cm}|p{1cm}p{1.5cm}p{1.5cm}}
\hline
& \multicolumn{3}{c}{Avareage Score}\\
\hline
Model & Baseline model & I.K. PresSim & Volumetric PresSim \\
\hline
MAE & 19.7 & 10.6 & 10.4 \\
Mask RMSD & 7.44 & 3.54 & 3.52 \\
Corrected $R^2$ & 0.407 & 0.800 & 0.811\\
Binarized $R^2$ & 0.659 & 0.909 & 0.948\\
\hline
\end{tabular}
}}
\end{center}
\caption{ Comparison table for average estimated score for baseline model and PresSim trained on deformation profiles simulated with inverse kinematics and Volumetric models respectively.}
\vspace{-20pt}
\label{table3}
\end{table}

\subsection{Faulty Cases}

One faulty case we encountered was where the point of contact between the limbs and the ground was entirely wrong.
This is not sporadic but continuous across the prediction sequence and can be observed while the subjects perform a defined set of yoga positions across different data collection sessions.
Any yoga sequence that involves extreme twisting/bending part of a body leads to an inaccurate humanoid rigged sequence and failure to predict the ground contact while simulating the deformations correctly. 
We assumed the most probable cause lies with using inverse kinematics for translating pose to animated sequence as it is not present in the PresSim model we trained using our volumetric approach.

Another failure case is where PresSim correctly predicts most parts of the shape of the point of contact between the body and the ground. However, the values of most pressure points are different, while some limb impressions are missing from the synthesized pressure maps.
Because of its sporadic nature across the prediction sequence, we assumed that the incorrect pose estimation of certain joints at a frame leads to this kind of error which can be observed in the PresSim model trained by either of our approaches.
 It can be solved using a more robust pose estimation model like multi-camera pose estimation or monocular pose estimation from two angles to synthesize deformation.

\section{Conclusion}
\label{co}
In conclusion, we present a framework combining video to 3D activities, physics simulations, and deep learning data transformation to synthesize sensor data of dynamic pressure profiles from monocular videos of human activities.
We evaluated our framework by comparing the floor pressure data generated by PresSim, a baseline model, and the ground truth sensor data collected using a smart fitness mat. 
To compare the shapes of the synthesized data with the ground truth, the \(R^2\) score calculated over binarized pressure maps across the dataset is 0.948, demonstrating high similarity.
We further demonstrate that on the individual sensitive node level, using a mask that considers only the point of contact between the body and mat, the corrected \(R^2\) score between the ground truth and synthesized pressure maps is, on average, 0.811.
This work introduces a sensing modality different from traditional pressure sensing devices, opening up new opportunities for estimating human body dynamics unaffected by limitations faced by contemporary pressure sensing devices and activity recognition, with potential applications in smart homes, healthcare, and rehabilitation.
 \section*{Acknowledgment}
 The research reported in this paper was supported by the BMBF (German Federal Ministry of Education and Research) in the project VidGenSense (01IW21003). 

\bibliographystyle{IEEEtran}
\bibliography{refs}

\end{document}